\documentclass[11pt,a4paper]{article}
\usepackage[hyperref]{acl2020}
\usepackage{times}
\usepackage{latexsym}
\usepackage{subfigure}
\usepackage{helvet} 
\usepackage{courier}  
\usepackage{graphicx}  

\usepackage{CJKutf8}
\usepackage{amsfonts,amssymb}
\usepackage{mathrsfs}
\usepackage{stfloats}
\usepackage{amsmath}
\usepackage{bm}
\usepackage{algorithm}
\usepackage{array}
\usepackage{mathrsfs}
\usepackage{algorithmic}
\usepackage{hhline}
\usepackage{multirow}
\usepackage{mathrsfs}
\usepackage[lined,boxed,commentsnumbered,algo2e]{algorithm2e}

\usepackage{microtype}
\aclfinalcopy 
\def\aclpaperid{1466} 
\title{Coupling Distant Annotation and Adversarial Training for Cross-Domain Chinese Word Segmentation}

\author{Ning Ding$^{1,2} $\hspace{0.5em}, Dingkun Long$^{2}$,  Guangwei Xu$^{2}$,\\ \textbf{Muhua Zhu}$^{2}$,  \textbf{Pengjun Xie}$^{2}$,  \textbf{Xiaobin Wang}$^{2}$, \textbf{Hai-Tao Zheng}$^{1}$\thanks{\quad Corresponding author}\hspace{0.5em} \\
$^{1}$Tsinghua University, China 
$^{2}$Alibaba Group\\
\texttt{\{dingn18\}@mails.tsinghua.edu.cn}, \texttt{\{zhumuhua\}@gmail.com},\\
\texttt{\{dingkun.ldk,kunka.xgw\}@alibaba-inc.com},\\
\texttt{\{chengchen.xpj,xuanjie.wxb\}@alibaba-inc.com},\\
\texttt{\{zheng.haitao\}@sz.tsinghua.edu.cn}\\ 
}

\date{}

\begin{document}
\maketitle
\begin{abstract}
Fully supervised neural approaches have achieved significant progress in the task of Chinese word segmentation (CWS). Nevertheless, the performance of supervised models tends to drop dramatically when they are applied to out-of-domain data. Performance degradation is caused by the distribution gap across domains and the out of vocabulary (OOV) problem. In order to simultaneously alleviate these two issues, this paper proposes to couple distant annotation and adversarial training for cross-domain CWS. For distant annotation, we rethink the essence of ``Chinese words'' and design an automatic distant annotation mechanism that does not need any supervision or pre-defined dictionaries from the target domain. The approach could effectively explore domain-specific words and distantly annotate the raw texts for the target domain. For adversarial training, we develop a sentence-level training procedure to perform noise reduction and maximum utilization of the source domain information. Experiments on multiple real-world datasets across various domains show the superiority and robustness of our model, significantly outperforming previous state-of-the-art cross-domain CWS methods. 

\end{abstract}

\section{Introduction}
\label{introduction}
Chinese is an ideographic language and lacks word delimiters between words in written sentences. Therefore, Chinese word segmentation (CWS) is often regarded as a prerequisite to downstream tasks in Chinese natural language processing~\cite{zhang2018chinese, li2019chinese, ding2019event}. This task is conventionally formalized as a character-based sequence tagging problem~\citep{peng2004chinese}, where each character is assigned a specific label to denote the position of the character in a word. With the development of deep learning techniques, recent years have also seen increasing interest in applying neural network models onto CWS~\citep{cai2016neural,liu2016exploring,cai2017fast,ma2018state}. These approaches have achieved significant progress on in-domain CWS tasks, but they still suffer from the cross-domain issue when they come to processing of out-of-domain data.

\begin{figure}[t]
\centering
\includegraphics[width=0.45\textwidth]{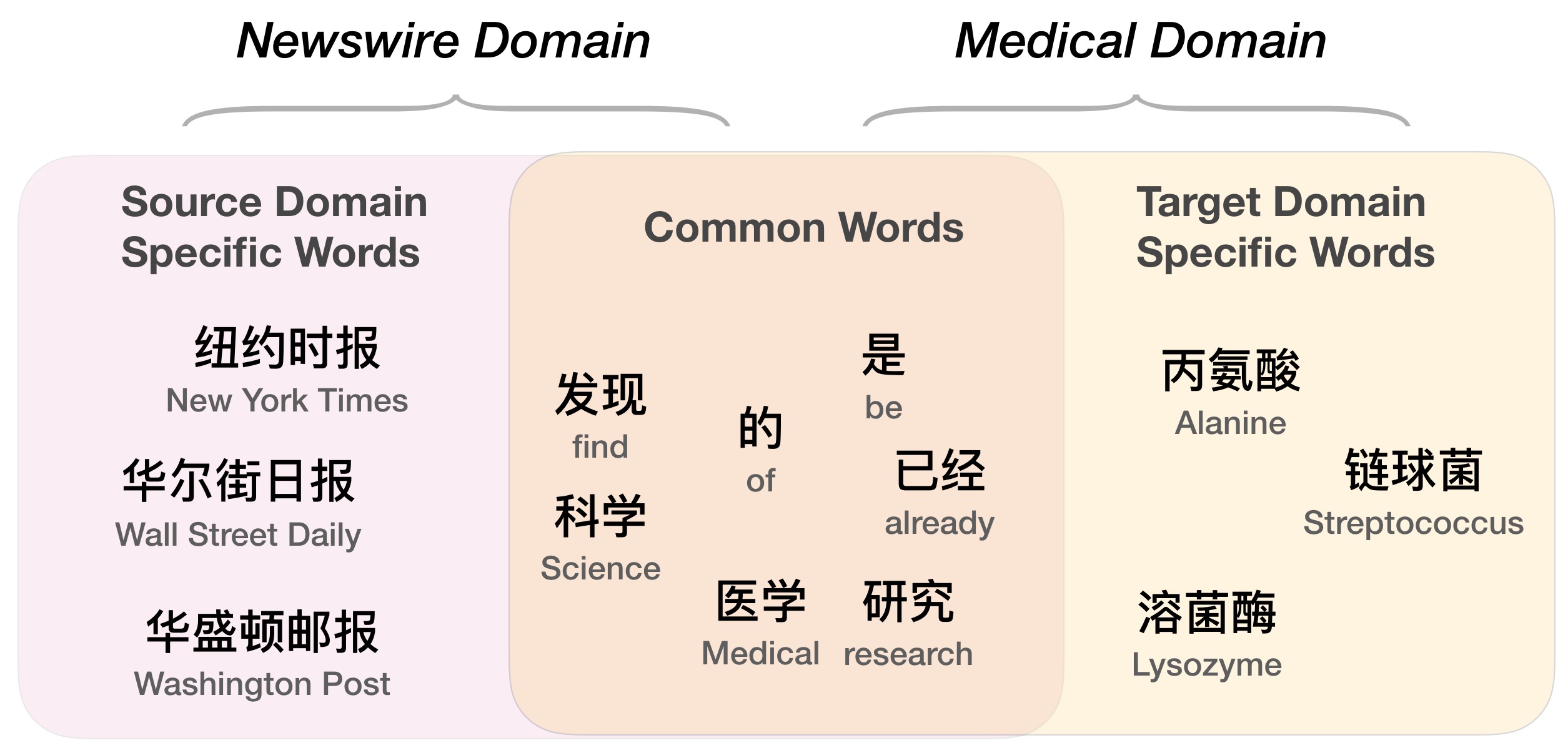} 
\vspace{-0.1cm}
\caption{Different word distributions for the newswire domain and the medical domain.}
\label{fig:1}
\vspace{-0.3cm}
\end{figure}

Cross-domain CWS is exposed to two major challenges: 1) \textbf{Gap of domain distributions}. This is a common issue existing in all domain adaptation tasks. Source domain data and target domain data generally have different distributions. As a result, models built on source domain data tend to degrade performance when they are applied to target domain data. Generally, we need some labeled target domain data to adapt source domain models, but it is expensive and time consuming to manually craft such data. 2) \textbf{Out of vocabulary (OOV) problem}, which means there exist some words in the testing data that never occur in the training data. Source domain models have difficulties in recognizing OOV words since source domain data contains no information on the OOVs. Figure~\ref{fig:1} presents examples to illustrate the difference between the word distributions of the newswire domain and the medical domain. Segmenters built on the newswire domain have very limited information to segment domain-specific words like ``\begin{CJK}{UTF8}{gbsn}溶菌酶\end{CJK} (Lysozyme)''. 

Previous approaches to cross-domain CWS mainly fall into two groups. The first group aims to attack the OOV issue by utilizing predefined dictionaries from the target domain to facilitate cross-domain CWS \cite{liu2014domain, zhao2018neural, zhang2018neural}, which are apt to suffer from scalability since not all domains possess predefined dictionaries. In other words, these methods are directly restricted by external resources that are available in a target domain. Studies in the second group \cite{ye2019improving} attend to learn target domain distributions like word embeddings from unlabeled target domain data. In this approach, source domain data is not fully utilized since the information from source domain data is transferred solely through the segmenter built on the data.

In this paper, we propose to attack the aforementioned challenges simultaneously by coupling the techniques of {\em distant annotation} and {\em adversarial training}. The goal of distant annotation is to automatically construct labeled target domain data with no requirement for human-curated domain-specific dictionaries. To this end, 
we rethink the definition and essence of ``Chinese words'' and develop a word miner to obtain domain-specific words from unlabeled target domain data. Moreover, a segmenter is trained on the source domain data to recognize the common words in unlabeled target data. This way, sentences from the target domain are assigned automatic annotations that can be used as target domain training data. 

Although distant annotation could provide satisfactory labeled target domain data, there still exist annotation errors that affect the final performance. To reduce the effect of noisy data in automatic annotations in target domain data and make better use of source domain data, we propose to apply adversarial training jointly on the source domain dataset and the distantly constructed target domain dataset. And the adversarial training module can capture deeper domain-specific and domain-agnostic features. 

To show the effectiveness and robustness of our approach, we conduct extensive experiments on five real-world datasets across various domains. Experimental results show that our approach achieves state-of-the-art results on all datasets, significantly outperforming representative previous works.  Further, we design sufficient subsidiary experiments to prove the alleviation of the aforementioned problems in cross-domain CWS. 

\section{Related Work}
\textbf{Chinese Word Segmentation} Chinese word segmentation is typically formalized as a sequence tagging problem. Thus, traditional machine learning models such as Hidden Markov Models (HMMs) and Conditional Random Fields (CRFs) are widely employed for CWS in the early stage \cite{wong1996chinese,gao2005chinese,zhao2010unified}. 
With the development of deep learning methods, research focus has been shifting towards deep neural networks that require little feature engineering.~\citet{chen2015long} are the first that use LSTM~\citep{hochreiter1997long} to resolve long dependencies in word segmentation problems. Since then, the majority of efforts is building end-to-end sequence tagging architectures, which significantly outperform the traditional approaches on CWS task \cite{wang2017convolutional,zhou2017word,yang2017neural,cai2017fast,chen2017adversarial,huang2019toward,gan2019investigating,yang2019subword}.

\noindent \textbf{Cross-domain CWS} As a more challenging task, cross-domain CWS has attracted increasing attention. \citet{liu2012unsupervised} propose an unsupervised model, in which they use a character clustering method and the self-training algorithm to jointly model CWS and POS-tagging. \citet{liu2014domain} apply partial CRF for cross-domain CWS via obtaining a partial annotation dataset from freely available data. Similarly, \citet{zhao2018neural} build partially labeled data by combining unlabeled data and lexicons. \citet{zhang2018neural} propose to incorporate the predefined domain dictionary into the training process via predefined handcrafted rules. \citet{ye2019improving} propose a semi-supervised approach that leverages word embeddings trained on the segmented text in the target domain. 
 
\noindent\textbf{Adversarial Learning} Adversarial learning is derived from the Generative Adversarial Nets (GAN) \cite{goodfellow2014generative}, which has achieved huge success in the computer vision field. Recently, many works have tried to apply adversarial learning to NLP tasks. \cite{jia2017adversarial,li2018generating,farag2018neural} focus on learning or creating adversarial rules or examples for improving the robustness of the NLP systems. For cross-domain or cross-lingual sequence tagging, the adversarial discriminator is widely used to extract domain or language invariant features~\citep{kim2017cross, huang2019cross, zhou2019dual}.


\begin{figure*}[t]
\setlength{\belowcaptionskip}{-10pt}
\centering
\includegraphics[width=1\textwidth]{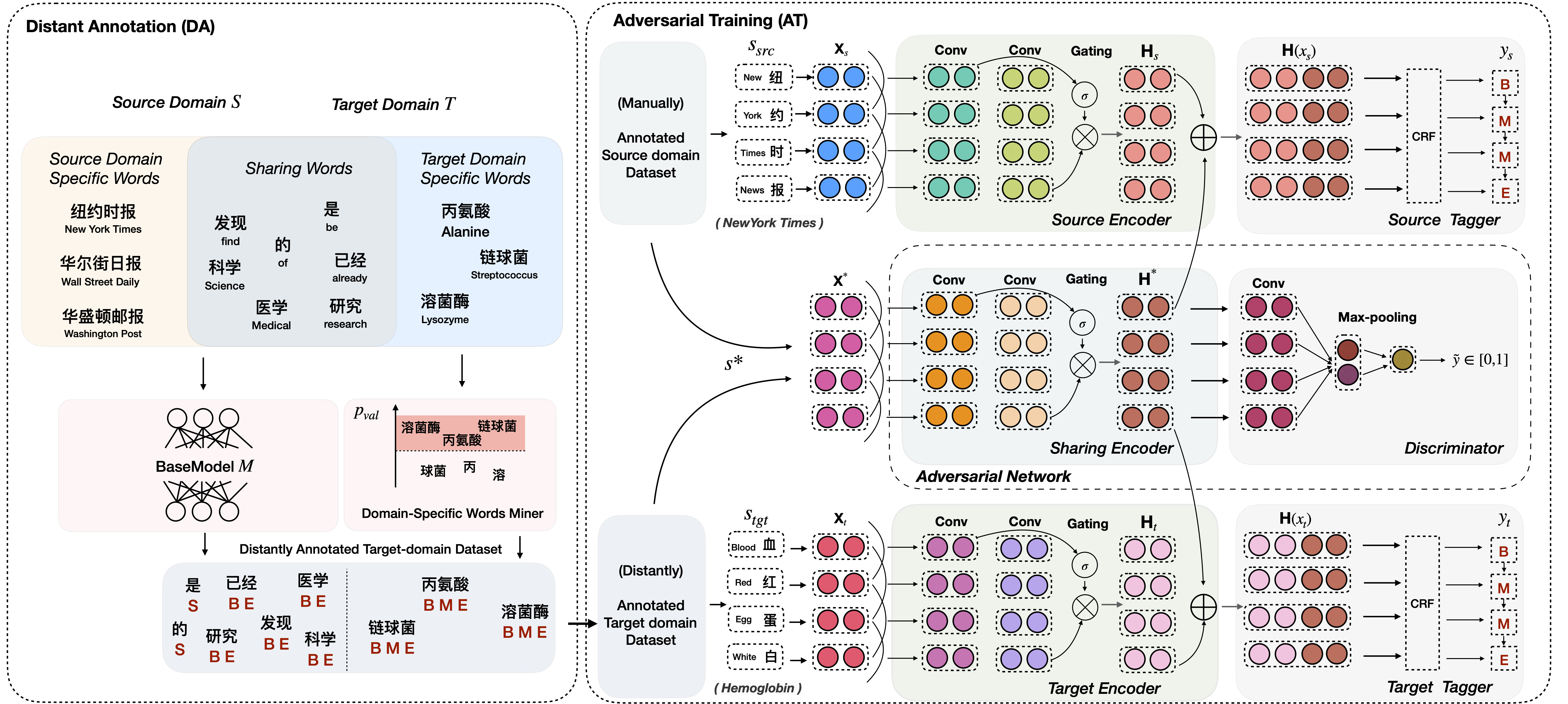} 
\caption{Detailed architecture of DAAT, the left part is the structure of the Distant Annotation (DA) module. The annotated dataset on target domain will be sent to the Adversarial Training (AT) module on the right part.}
\label{fig2}
\end{figure*}

\section{Our Approach}
\label{methodology}
Figure \ref{fig2} shows the framework of our approach to cross-domain CWS, which is mainly composed of two components: 1) \textbf{Distant Annotation (DA)}, and 2) \textbf{Adversarial Training (AT)}. In the following, we will describe details of the framework (DAAT) from the left to right in Figure \ref{fig2}.





In this paper, bold-face letters (e.g. $\bm{W}$) are used to denote vectors, matrices and tensors. We use numerical subscripts to indicate the indices of a sequence or vector. We use the subscript of $src$ to indicate the source domain and $tgt$ to denote the target domain. 

\subsection{Distant Annotation}
\label{sub-DA}
As illustrated in Figure \ref{fig2}, given a labeled source domain dataset and an unlabeled target domain dataset, distant annotation (DA) aims to automatically generate word segmentation results for sentences in the target domain. DA has two main modules, including a base segmenter and a \textit{Domain-specific Words Miner}. Specifically, the base segmenter is a GCNN-CRF \cite{wang2017convolutional} model trained solely on the labeled source domain data and is used to recognize words that are common among the source and target domains. \textit{Domain-specific Words Miner} is designed to explore the target domain-specific words. 

\noindent{\bf Base Segmenter} In the CWS task, given a sentence $s$ = $\{ c_1,c_2,...,c_n\}$ , following the {\it BMES} tagging scheme, each character $c_i$ is assigned one of the labels in $\{B,M,E,S\}$, indicating whether the character is in the beginning, middle, end of a word, or the character is merely a single-character word. 

For a sentence $s$, we first use an embedding layer to obtain the embedding representation $\bm{e}_i$ for each character $c_i$. Then, the sentence $s$ can be represented as $\bm{e}$ = $\{\bm{e}_1, \bm{e}_2, ...,\bm{e}_n\} \in \mathbb{R}^{n\times d}$, where $d$ denotes the embedding dimension. $\bm{e}$ will be fed into the GCNN model \cite{dauphin2017language,gehring2017convolutional}, which computes the output as:
\begin{equation}
\bm{H}_s = (\bm{e} * \bm{W} + b) \odot \sigma(\bm{e} * \bm{V} + c),
\end{equation}
here, $\bm{W} \in \mathbb{R}^{k \times d \times l}$, $b\in \mathbb{R}^l$, $\bm{V} \in \mathbb{R}^{k \times d \times l}$, $c \in \mathbb{R}^l$.  $d$ and $l$ are the input and output dimensions respectively, and $k$ is the window size of the convolution operator. $\sigma$ is the sigmoid function and $\odot$ represents element-wise product. We adopt a stacking convolution architecture to capture long distance information, the output of the previous layers will be treated as input of the next layer. The final representation of sentence $s$ is $ \bm{H}_s = \{\bm{h}_1,\bm{h}_2,...,\bm{h}_n\}$.

Correlations among labels are crucial factors in sequence tagging. Particularly, for an input sequence $s_{src} = \{c_1, c_2,..., c_n\}$ (take source domain data as example), the corresponding label sequence is $L = \{y_1, y_2, ..., y_n\}$. The goal of CRF is to compute the conditional probability distribution:
\begin{equation}
        P(L|s_{src})\! =\!\! \frac{\!\!exp(\sum\limits^n_{i=1}\!(S(y_i)\!\!+\!\!T(y_{i-1},y_i)))}{\sum\limits_{L'\in \mathbb{C}}\!\!\!exp(\sum\limits^n_{i=1}\!(S(y'_i)\!\!+\!\!T(y'_{i-1},y'_i)))},
\end{equation}
where $T$ denotes the transition function to calculate the transition scores from $y_{i-1}$ to $y_i$. $\mathbb{C}$ contains all the possible label sequences on sequence $s$ and $L'$ is a random label sequence in $\mathbb{C}$. And $S$ represents the score function to compute the emission score from the hidden feature vector $\bm{h}_i$ to the corresponding label $y_i$, which is defined as:
\begin{equation}
  S(y_i) = \bm{W}^{y_i}\bm{h}_i+b^{y_i},
\end{equation}
$\bm{W}^{y_i}$ and  $b^{y_i}$ are learned parameters specific to the label $y_i$. 

To decode the highest scored label sequence, a classic Viterbi \cite{viterbi1967error} algorithm is utilized as the decoder. The loss function of the sequence tagger is defined as the sentence-level negative log-likelihood:
\begin{equation}
   \mathcal{L}_{src}= - \sum{\rm log}\,P(L|s_{src}).
\end{equation}
The loss of the target tagger $\mathcal{L}_{tgt}$ could be computed similarly.

\noindent {\bf Domain-specific Words Miner} As mentioned in section \ref{introduction},  previous works usually use existing domain dictionaries to solve the domain-specific noun entities segmentation problem in cross-domain CWS. But this strategy does not consider that it is properly difficult to acquire a dictionary with high quality for a brand new domain. In contrast, we develop a simple and efficient strategy to perform domain-specific words mining without any predefined dictionaries.

Given large raw text on target domain and a base segmenter, we can obtain a set of segmented texts $\Gamma$ = $\{T_1, T_2, ...,T_N\}$, where stop-words are removed. Then let $\gamma = \{t_1,t_2,...,t_m \}$ denote all the n-gram sequences extracted from $\Gamma$. For each sequence $t_i$, we need to calculate the possibility that it is a valid word. In this procedure, four factors are mainly considered.

\begin{figure}[t]
\setlength{\belowcaptionskip}{-10pt}
\centering
\subfigure[Mutual Information to measure the internal tightness. ]{\includegraphics[width=0.45\textwidth]{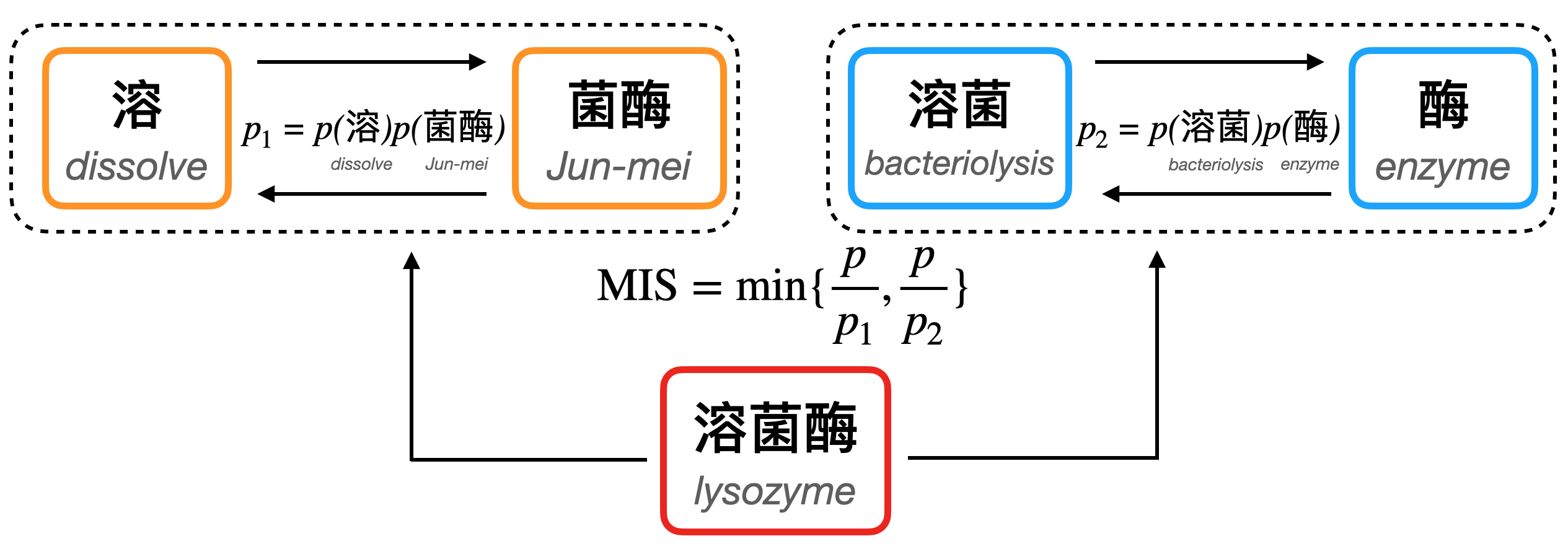}}
\subfigure[Entropy Score to measure the external flexibility.]{\includegraphics[width=0.45\textwidth]{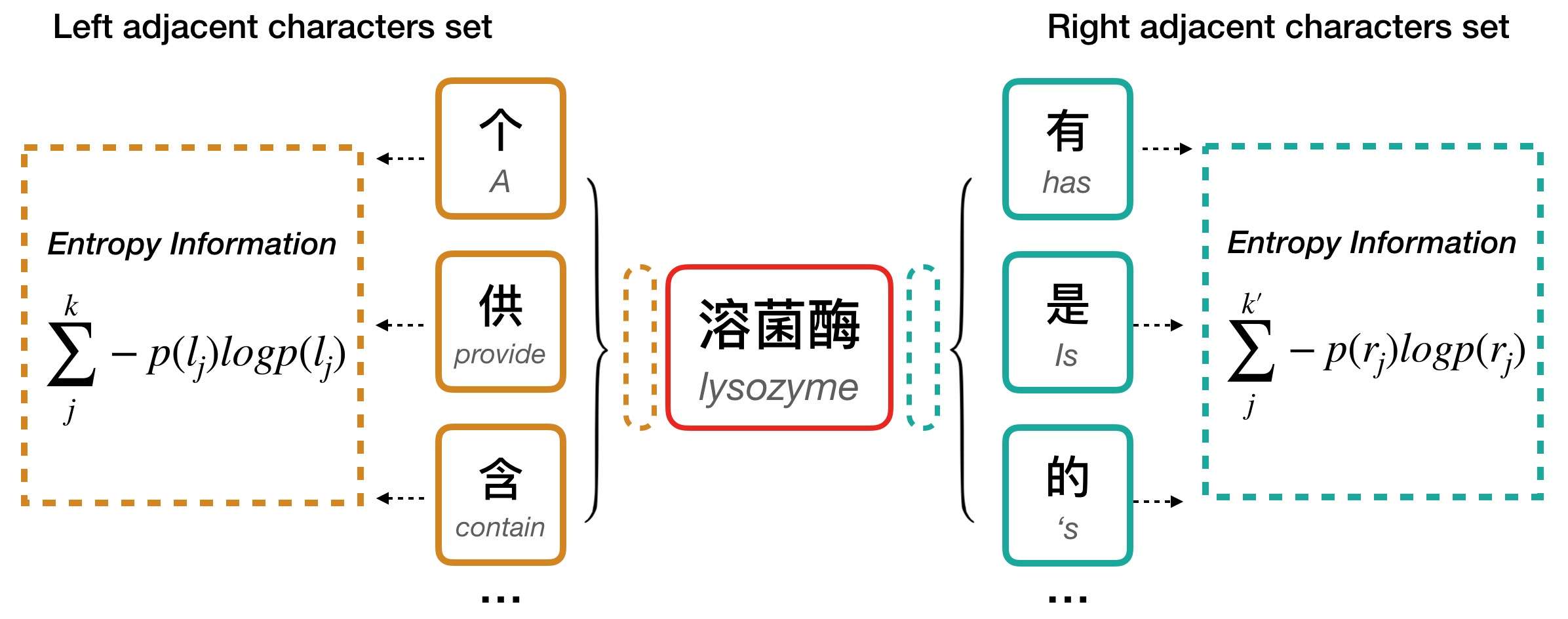}}
\caption{Examples of Mutual Score and Entropy Information factors.}.
\label{fig:cases}
\end{figure}

\noindent1) {\it Mutual Information} (MI). MI~\citep{kraskov2004estimating} is widely used to estimate the correlation of two random variables. Here, we use mutual information between different sub-strings to measure the internal tightness for a text segment, as shown in Figure \ref{fig:cases}(a).  
Further, in order to exclude extreme cases, it is necessary to enumerate all the sub-string candidates. The final $\rm MI$ score for one sequence $t_i$ consists of $n$ characters $t_i = \{c_1...c_n\}$ is defined as:
\begin{equation}
    {\rm MIS}(t_i)\! =\! \min_{j \in [1:n]}\{\frac{p(t_i)}{p(c_1...c_j)\cdot p(c_{j+1}...c_{n})}\},
\end{equation}
where $p(\cdot)$ denotes the probability given the whole corpus $\Gamma$.

\noindent2) {\it Entropy Score} (ES). Entropy is a crucial concept aiming at measuring the uncertainty of random variables in information theory~\citep{jaynes1957information}. Thus, we can use ES to measure the uncertainty of candidate text fragment, since higher uncertainty means a richer neighboring context. Let $N_l(t_i)=\{l_1,...,l_k\}$ and $ N_r(t_i)=\{r_1,...,r_{k'}\}$ be the set of left and right adjacent characters for $t_i$. The left entropy score ${\rm ES}_{l}$ and right entropy ${\rm ES}_{r}$ of $t_i$ can be formulated as ${\rm ES}_{l}(t_i) $=$\sum^k_j -p(l_j){\rm log}\, p(l_j)$ and  ${\rm ES}_{r}(t_i)$=$\sum^{k'}_j -p(r_j){\rm log}\, p(r_j)$ respectively. We choose ${\min}({\rm ES}_{l}(t_i), {\rm ES}_{r}(t_i))$ as the final score for $t_i$. Hence, ${\rm ES}(t_i)$ could explicitly represent the external flexibility for a text segment (as shown in Figure~\ref{fig:cases}(b)), and further serve as an important indicator to judge whether the segment is an independent word. 

\noindent3) {\it tf-idf}. tf-idf is a widely used numerical statistic that can reflect how important a word is to a document in a collection or corpus. As illustrated in Figure \ref{fig:1},  most of the domain-specific words are noun entities, which share a large weighting factor in general.

In this work, we define a word probability score $p_{val}(t_i)$ to indicate how likely $t_i$ can be defined as a valid word. 
\begin{equation}
p_{val}(t_i)\!\!=\!\sigma({\rm N}[{\rm MIS}(t_i)] + {\rm N}[{\rm ES}(t_i)] + {\rm N}[{\rm tfidf}(t_i)]),
\end{equation}
where $\sigma$ denotes the {\rm sigmoid} function and $\rm N$ denotes normalization operation with the max-min method. 

\noindent4) {\it Word frequency}. If $t_i$ is a valid word, it should appear repeatedly in $\Gamma$. 

Finally, by setting an appropriate threshold for $p_{val}(t_i)$ and word frequence, the \textit{Domain-Specific Words Mine}r could effectively explore domain-specific words, then construct the domain-specific word collection $\mathcal{C}$ for the target domain. In this work, we only consider words $t_i$ with $p_{val}(t_i) \geq 0.95$ and frequency larger than $10$.

The left part of Figure \ref{fig2} illustrates the data construction process of \textit{DA}.  First, we utilize the \textit{Domain-specific Words Miner} to build the collection $\mathcal{C}$ for the target domain. Take sentence ``\begin{CJK}{UTF8}{gbsn}溶酶菌的科学研究\end{CJK} (Scientific research on lysozyme)'' as an example, we use the forward maximizing match algorithm based on $\mathcal{C}$, which shows that ``\begin{CJK}{UTF8}{gbsn}溶酶菌\end{CJK} (lysozyme)" is a valid word. Hence, the labels of characters ``\begin{CJK}{UTF8}{gbsn}溶\end{CJK}", ``\begin{CJK}{UTF8}{gbsn}酶\end{CJK}", ``\begin{CJK}{UTF8}{gbsn}菌\end{CJK}" are ``$B$", ``$M$", ``$E$". For the left part of the sentence, we adopt the baseline segmenter to perform the labelling process. ``\begin{CJK}{UTF8}{gbsn}的科学研究\end{CJK}" will be assigned with $\{``S",``B".``E",``B",``E"\}$. To this end, we are able to automatically build annotated dataset on the target domain.

\subsection{Adversarial Training}
\label{sub:adv}
The structure of the Adversarial Training module is illustrated as the right part of Figure \ref{fig2}. As mentioned in \ref{sub-DA}, we construct an annotated dataset for the target domain. Accordingly, the inputs of the network are two labeled datasets from source domain $\mathcal{S}$ and target domain $\mathcal{T}$.  There are three encoders to extract features with different emphases, and all the encoders are based on GCNN as introduced in section \ref{sub-DA}. For domain-specific features, we adopt two independent encoders $E_{src}$ and $E_{tgt}$ for source domain and target domain. For domain-agnostic features, we adopt a sharing encoder $E_{shr}$ and a discriminator $G_d$, which will be both trained as adversarial players. 

For the two domain-specific encoders, the input sentence is $s_{src}$=$\{c^s_1, c^s_2,...,c^s_n\}$ from source domain, or sentence $s_{tgt}$=$\{c^t_1, c^t_2,...,c^t_m\}$ from the target domain. The sequence representation of $s_{src}$ and $s_{tgt}$ can be obtained by $E_{src}$ and $E_{tgt}$. Thus, the domain independent representations of $s_{src}$ and $s_{tgt}$ are $\bm{H}_{s} \in \mathbb{R}^{n\times l}$ and $ \bm{H}_{t} \in \mathbb{R}^{m\times l}$, where $n$ and $m$ denote the sequence lengths of $s_{src}$ and $s_{tgt}$ respectively, $l$ is the output dimension of GCNN encoder.

For the sharing encoder, we hope that $E_{shr}$ is able to generate representations that could fool the sentence level discriminator to correctly predict the domain of each sentence, such that $E_{shr}$ finally extracts domain-agnostic features. Formally, given sentences $s_{src}$ and $s_{tgt}$ from source domain and target domain, $E_{shr}$ will produce sequence features $\bm{H}_s^*$ and $\bm{H}_t^*$ for $s_{src}$ and $s_{tgt}$ respectively.

The discriminator ${G_d}$ of the network aims to distinguish the domain of each sentence. Specifically, we will feed the final representation $\bm{H}^*$ of every sentence $s$ to a binary classifier $G_y$ where we adopt the text CNN network \cite{kim2014convolutional}. $G_y$ will produce a probability that the input sentence $s$ is from the source domain or target domain.
Thus, the loss function of the discriminator is:
\begin{equation}
    \begin{split}
    \mathcal{L}_{d} = & -\mathbb{E}_{s \sim p_{\mathcal{S}}(s)} [{\rm log}\, G_y(E_{shr}(s)]  \\ & -\mathbb{E}_{s \sim p_{\mathcal{T}}(s)} [{\rm log}\,
    (1-G_y(E_{shr}(s))],
    \end{split}
\end{equation}
Features generated by the sharing encoder $E_{shr}$ should be able to fool the discriminator to correctly predict the domain of $s$. Thus, the loss function for the sharing encoder $\mathcal{L}_c$ is a flipped version of $\mathcal{L}_d$:
\begin{equation}
    \begin{split}
    \mathcal{L}_{c} = & -\mathbb{E}_{s \sim p_{\mathcal{S}}(s)} [{\rm log}\ (1- G_y(E_{shr}(s)])  \\ & -\mathbb{E}_{s \sim p_{\mathcal{T}}(s)} [{\rm log} G_y(E_{shr}(s)],
    \end{split}
\end{equation}

Finally, we concatenate $\bm{H}$ and $\bm{H}^*$ as the final sequence representation of the input sentence. For $s_{src}$ from source domain, $\bm{H}(s_{src}) = [ \bm{H}_s \oplus \bm{H}_s^*]$, while for $s_{tgt}$ from the target domain, $\bm{H}(s_{tgt}) = [ \bm{H}_t \oplus \bm{H}_t^*]$. The final representation will be fed into the CRF tagger.

So far, our model can be jointly trained in an end-to-end manner with the standard back-propagation algorithm. More details about the adversarial training process are described in Algorithm 1. When there is no annotated dataset on the target domain, we could remove $\mathcal{L}_{tgt}$ during the adversarial training process and use the segmenter on source domain for evaluation.







\begin{algorithm}[H]
  \SetAlgoLined
  \textbf{Input: }{Manually annotated dataset $\mathcal{D}_s$ for source domain $\mathcal{S}$, and distantly annotated dataset $\mathcal{D}_t$} for target domain $\mathcal{T}$
  
  \For{$i\leftarrow 1$ \KwTo epochs}{
      \For{$j\leftarrow 1$ \KwTo num\_of\_steps per epoch}{
        Sample mini-batches $\mathcal{X}_s \sim \mathcal{D}_s$, $\mathcal{X}_t \sim \mathcal{D}_t$ \;
        
        \eIf{$j \% 2=1$}{
            $loss$ = $\mathcal{L}_{src}$ + $\mathcal{L}_{tgt}$ + $\mathcal{L}_d$\;
            
            Update $\theta$ w.r.t $loss$ \;
            }{
            $loss$= $\mathcal{L}_{src}$ + $\mathcal{L}_{tgt}$ + $\mathcal{L}_c$\;
            
            Update $\theta$ w.r.t $loss$ \;
            }

        }
  }
  \label{adv_train}
  \caption{Adversarial training algorithm.}
\end{algorithm}

\section{Experiments}
In this section, we conduct extensive cross-domain CWS experiments on multiple real-world datasets with different domains, then comprehensively evaluate our method and other approaches.


\subsection{Datasets and Experimental Settings}
\begin{table}[]
\setlength{\belowcaptionskip}{-10pt}
\centering
\scalebox{0.78}{
\begin{tabular}{c|c|c|ccc|c}
\hline
\multicolumn{3}{c|}{\textbf{Dataset}}                                  & \textbf{Sents}          & \textbf{Words}         & \textbf{Chars}          & \textbf{Domain}                                                                     \\ \hline
\multirow{2}{*}{SRC}  & \multirow{2}{*}{PKU} & Train          & 47.3K          & 1.1M           & 1.8M           & \multirow{2}{*}{News}                                                      \\
                      &                      & Test           & 6.4K           & 0.2M           & 0.3M           &                                                                            \\\hhline{=======}
\multirow{10}{*}{TGT} & \multirow{2}{*}{DL}  & Full          & 40.0K          & 2.0M           & 2.9M           & \multirow{2}{*}{\begin{tabular}[c]{@{}c@{}}Novel\end{tabular}} \\
                      &                      & Test           & 1.0K           & 32.0K          & 47.0K          &                                                                            \\ \cline{2-7} 
                      & \multirow{2}{*}{FR}  & Full          & 148K           & 5.0M           & 7.1M           & \multirow{2}{*}{\begin{tabular}[c]{@{}c@{}}Novel \end{tabular}} \\
                      &                      & Test           & 1.0K           & 17.0K          & 25.0K          &                                                                            \\ \cline{2-7} 
                      & \multirow{2}{*}{ZX}  & Full          & 59.0K          & 2.1M           & 3.0M           & \multirow{2}{*}{\begin{tabular}[c]{@{}c@{}}Novel \end{tabular}} \\
                      &                      & Test           & 1.0K           & 21K            & 31.0K          &                                                                            \\ \cline{2-7} 
                      & \multirow{2}{*}{DM}  & Full          & 32.0K          & 0.7M           & 1.2M           & \multirow{2}{*}{Medical}                                                   \\
                      &                      & Test           & 1.0K           & 17K            & 30K            &                                                                            \\ \cline{2-7} 
                      & \multirow{2}{*}{PT}  & Full & 17.0K & 0.6M  & 0.9M  & {\multirow{2}{*}{Patent}}                      \\
                      &                      & Test & 1.0K  & 34.0K & 57.0K &                                                   \\ \hline
\end{tabular}}
\caption{Statistics of datasets. The datasets of the target domain (TGT) are originally raw texts without golden segmentation, and the statistics are obtained by the baseline segmenter. The \textit{DA} module will distantly annotate the datasets as mentioned in \ref{sub-DA}.}
\label{table:dataset}
\end{table}

\noindent\textbf{Datasets} Six datasets across various domains are used in our work. The statistics of all datasets are shown in Table \ref{table:dataset}. In this paper, we use PKU dataset \cite{emerson2005second} as the source domain data, which is a benchmark CWS dataset on the newswire domain. In addition, the other five datasets in other domains will be utilized as the target domain datasets. Among the five target domain datasets there are three Chinese fantasy novel datasets, including DL \textit{(DoLuoDaLu)}, FR \textit{(FanRenXiuXianZhuan)} and ZX \textit{(ZhuXian)} \cite{qiu2015word}. An obvious advantage for fantasy novel datasets is that there are a large number of proper words originated by the author for each fiction, which could explicitly reflect the alleviation of the OOV problem for an approach. Besides the fiction datasets, we also use DM (dermatology) and PT (patent) datasets \cite{ye2019improving}, which are from \textit{dermatology} domain and \textit{patent} domain respectively. All the domains of the target datasets are very different from the source dataset (newswire). To perform a fair and comprehensive evaluation, the full/test settings of the datasets follow \citet{ye2019improving}.

\noindent\textbf{Hyper-Parameters} Table \ref{table:hyperp} shows the hyper-parameters used in our method. All the models are implemented with Tensorflow \cite{abadi2016tensorflow} and trained using mini-batched back-propagation. Adam optimizer \cite{KingmaB14} is used for optimization. The models are trained on NVIDIA Tesla V100 GPUs with CUDA. \footnote{source code and dataset will be available at \url{https://
	github.com/Alibaba-NLP/DAAT-CWS}}
 
\noindent\textbf{Evaluation Metrics} We use standard micro-averaged precision (P), recall (R) and F-measure as our evaluation metrics. We also compute OOV rates to reflect the degree of the OOV issue.

\subsection{Compared Methods}
We make comprehensive experiments with selective previous proposed methods, which are: 
\textbf{Partial CRF} \cite{liu2014domain} builds partially annotated data using raw text and lexicons via handcrafted rules, then trains the CWS model based on both labeled dataset (PKU) and partially annotated data using CRF.  
\textbf{CWS-DICT} \cite{zhang2018neural} trains the CWS model with a BiLSTM-CRF architecture, which incorporates lexicon into a neural network by designing handcrafted feature templates. 
For fair comparison, we use the same domain dictionaries produced by the \textit{Domain-specific Words Miner} for {\bf Partial CRF} and {\bf CWS-DICT} methods. 
\textbf{WEB-CWS} \cite{ye2019improving} is a semi-supervised word-based approach using word embeddings trained with segmented text on target domain to improve cross-domain CWS. 

Besides, we implement strong baselines to perform a comprehensive evaluation, which are:
\textbf{GCNN (PKU)} uses the PKU dataset only, and we adopt the GCNN-CRF sequence tagging architecture \cite{wang2017convolutional}.  
\textbf{GCNN (Target)} uses the distantly annotated dataset built on the target domain only.
\textbf{GCNN (Mix)} uses the mixture dataset with both the PKU dataset and the distantly annotated target domain dataset.
\textbf{DA} is a combination of GCNN (PKU) and domain-specific words. Details are introduced in \ref{sub-DA}.
\textbf{AT} denotes the setting that we adopt adversarial training when no distantly annotated dataset on the target domain is provided, but the raw text is available. 

\subsection{Overall Results}
\begin{table}[t]
\setlength{\belowcaptionskip}{-10pt}
\centering
\scalebox{0.84}{
\setlength{\tabcolsep}{26 pt}{
\begin{tabular}{lc}
\hline
\textbf{Hyper-parameter Name} & \textbf{Value} \\ \hline
    Threshold for $p_{val}$ & 0.95 \\
    Char emb size  &   200    \\
    GCNN output dim    &  200     \\
    Text CNN num of filters & 200 \\
    Text CNN filter size & [3,4,5] \\
    GCNN layers             &    5   \\
    Dropout Rate             &   0.3    \\
    Batch size             &    128   \\
    Learning rate      &  0.001 \\
    Epochs             &    30  \\ 
    \hline
\end{tabular}}}
\caption{Hyper-parameters.}
\label{table:hyperp}
\end{table}

    
    
    


\begin{table*}[tp]
\setlength{\belowcaptionskip}{-10pt}
\centering
\scalebox{0.73}{
\begin{tabular}{c|ccc||cccccc}
\hhline{----||------}
\multirow{2}{*}{\textbf{Dataset}} & \multicolumn{3}{c||}{\textbf{Previous Methods (F1-score)}} & \multicolumn{5}{c}{\textbf{Ours (F1-score)}}                                 \\  \hhline{~---||------} 
                        & Partial CRF     & CWS-DICT     & WEB-CWS  & AT   & GCNN (PKU) & DA  & GCNN(Mix) &  GCNN (Target) & \textbf{DAAT} \\ \hhline{----||------}
DL                      &92.5                 & 92.0         & \textbf{93.5}    &90.7    & 90.0     &   93.6      &   93.9   &  93.9           & \textbf{94.1 (+0.6)}  \\
FR                      &\textbf{90.2}                 & 89.1         & 89.6    &86.8    & 86.0     &   92.4      &   92.6   & 92.6           & \textbf{93.1 (+2.9)}  \\
ZX                      &83.9                 & 88.8          & \textbf{89.6}    &85.0    & 85.4     &   90.4      &   90.6    &  90.7           & \textbf{90.9 (+1.3)}  \\
DM                      &\textbf{82.8}                 & 81.2          & 82.2    &81.0    & 82.4     &   83.8      &   83.9    &84.3             & \textbf{85.0 (+2.2)}  \\
PT                      &85.0                 &  \textbf{85.9}        & 85.1    &85.1    & 87.6     & 89.1        &   89.3    & 89.3            & \textbf{89.6 (+3.7)}  \\ \hhline{----||------}
\end{tabular}
}
\caption{The overall results on five datasets. The first block contains the latest cross-domain methods. And the second block reports the results for our implemented methods and DAAT. Numbers in the parentheses indicate absolute improvement than previous SOTA results.}
\label{table:finalresults} 
\end{table*}

\begin{table}[tp]
\setlength{\belowcaptionskip}{-10pt}
\centering
\scalebox{0.78}{
\begin{tabular}{cccc}
\hline
\multicolumn{2}{c|}{\textbf{Dataset}}                                             & \textbf{OOV rate (source)} & \textbf{OOV rate (target)} \\ \hline
\multicolumn{1}{c|}{Source}                  & \multicolumn{1}{c|}{PKU} & 3.70\%          & -               \\ \hhline{====}
\multicolumn{1}{c|}{\multirow{5}{*}{Target}} & \multicolumn{1}{c|}{DL}  & 11.15\%         & \textbf{1.23\%}          \\
\multicolumn{1}{c|}{}                        & \multicolumn{1}{c|}{FR}  & 14.08\%         & \textbf{0.98\%}          \\
\multicolumn{1}{c|}{}                        & \multicolumn{1}{c|}{ZX}  & 15.52\%         & \textbf{1.43\%}          \\
\multicolumn{1}{c|}{}                        & \multicolumn{1}{c|}{DM}  & 25.93\%         & \textbf{5.42\%}          \\
\multicolumn{1}{c|}{}                        & \multicolumn{1}{c|}{PT}  & 18.39\%         & \textbf{3.45\%}          \\ \hline
\end{tabular}}
\vspace{-0.2cm}
\caption{OOV rates on five datasets. OOV rate (source) means the OOV rate test dataset and PKU dataset. OOV rate (target) means the OOV rate between the test dataset and the constructed annotated target dataset.}
\label{table:oov1}
\end{table}


The final results are reported in Table \ref{table:finalresults}, from which we can observe that:

(1) Our DAAT model significantly outperforms previously proposed methods on all datasets, yielding the state-of-the-art results. Particularly, DAAT improves the F1-score on the five datasets from 93.5 to 94.1, 90.2 to 93.1, 89.6 to 90.9, 82.8 to 85.0 and 85.9 to 89.6 respectively.  The results demonstrate that the unified framework is empirically effective, for the alleviation of the OOV problem and the full utilization of source domain information.

(2) As mentioned in section \ref{methodology}, the AT model uses the same adversarial training network as the DAAT, yet without annotation on the target domain dataset. Results on the AT setting could explicitly reflect the necessity to construct the annotated target domain dataset. Specifically, without the constructed dataset, the AT method only yields 90.7, 86.8, 85.0, 81.0 and 85.1 F1-scores on five datasets respectively. But when use the annotated target domain dataset, we can get the DAAT with the best performance. 

(3) WEB-CWS was the state-of-the-art approach that utilizes word embeddings trained on the segmented target text. Yet it is worth noticing that our model that only combines the base segmenter trained on PKU and domain-specific words (\textit{DA}) could outperform WEB-CWS, which indicates that the distant annotation method could exploit more and deeper semantic features from the raw text. For the CWS-DICT method, which requires an external dictionary, we use the word collection (built by the \textit{Domain-specific Words Miner}) to guarantee the fairness of the experiments. We can observe that our framework could yield significantly better results than CWS-DICT. Moreover, CWS-DICT needs existing dictionaries as external information, which is difficult for the model to transfer to brand new domains without specific dictionaries. In contrast, our framework utilizes the \textit{Domain-specific Words Miner} to construct the word collection with high flexibility across domains.

\subsection{Effect of Distant Annotation}
In this section, we focus on exploring the ability to tackle the \textbf{OOV problem} for the \textit{DA} method, which could distantly construct an annotated dataset from the raw text on the target domain. As illustrated in Table \ref{table:oov1}, the cross-domain CWS task suffers from a surprisingly serious OOV problem. All OOV rates (source) are above 10\%, which will definitely degrade model performance. Nevertheless, after constructing an annotated dataset on the target domain, the OOV rate (target) drops significantly. Specifically, the \textit{DA} method yields 9.92\%, 13.1\%, 14.09\% 20.51\% and 14.94\% absolute OOV rate drop on the five out-domain datasets. The statistical result reveals that the \textit{Domain-specific Words Miner} could accurately explore specific domain words for any domains from raw texts. Therefore, the \textit{DA} of our framework could efficaciously tackle the OOV problem. Moreover, the module does not need any specific domain dictionaries,  which means it can be transferred to new domains without limitations.
\subsection{Impact of the Threshold $p_{val}$ }
\begin{figure}[h]
\centering
\includegraphics[width=0.48\textwidth]{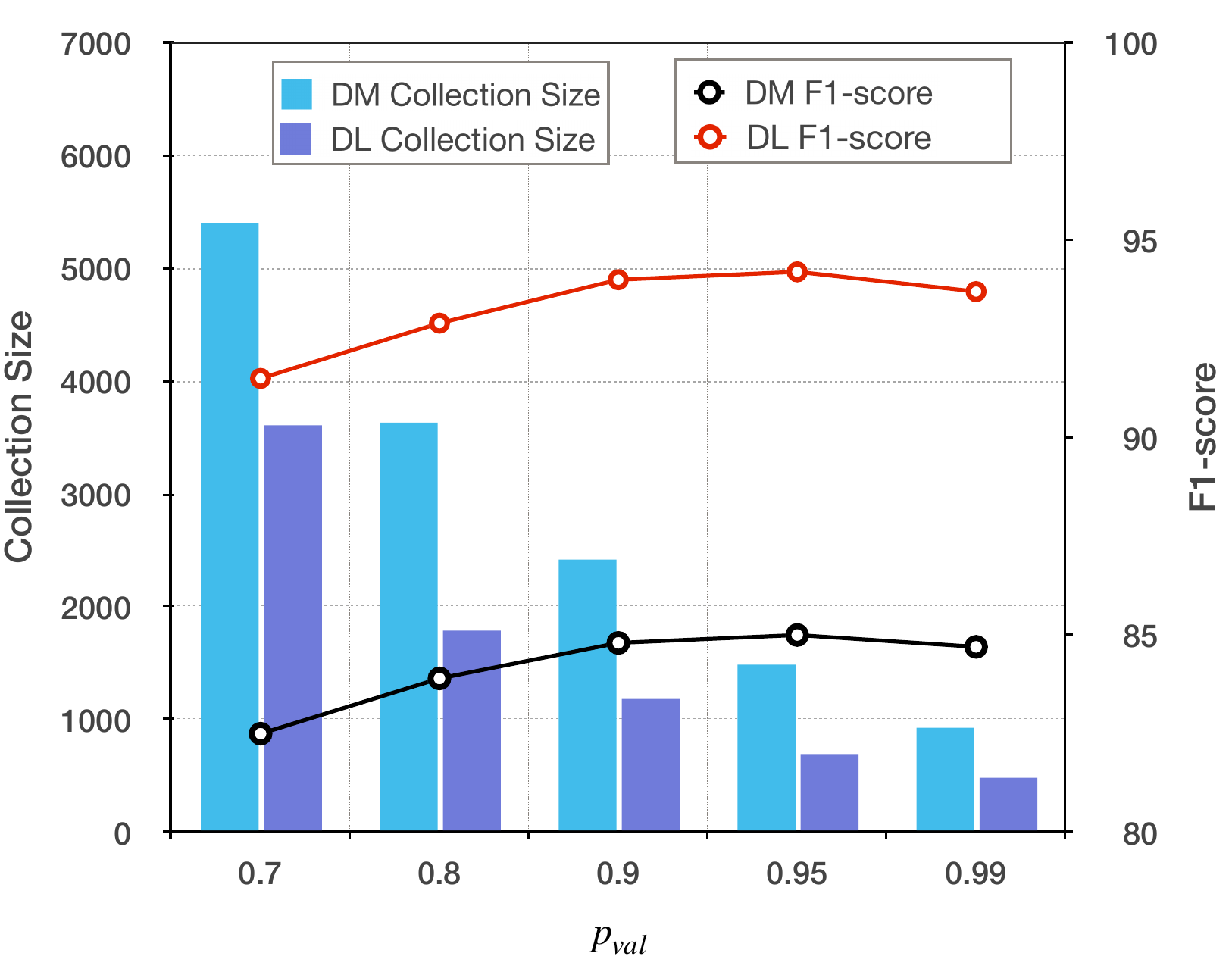} 
\caption{The impact of different $p_{val}$ on mined collection size and model performance.}
\label{fig:pval}
\end{figure}

Obviously, the setting of the hyper-parameter $p_{val}$ will directly affect the scale and quality of the domain-specific word collection. To analyze how $p_{val}$ affects the model performance, we conduct experiments with different setting $p_{val}$ in $\{0.7, 0.8, 0.9, 0.95, 0.99\}$, and the size of word collection and model performance on DL and DM datasets are shown in Figure \ref{fig:pval}. Constant with intuition, the collection size will decrease as the increase of $p_{val}$ because the filter criterion for words will get more strict, which is also a process of noise reduction. However, the F1-score curves are not incremental or descending. When $p_{val} <= 0.95$, the F1-scores on two datasets will increase because the eliminated words of this stage are mostly wrong. While the F1-scores will maintain or decrease when $p_{val} > 0.95$, because in this case, some correct words will be eliminated. We set $p_{val} = 0.95$ to guarantee the quality and quantity of the word collection simultaneously, so as to guarantee the model performance. And in this setting, the collection sizes are 0.7k words for DL, 1.7k for FR, 3.3k for ZX, 1.5k for DM and 2.2k for PT respectively.

\setlength{\belowcaptionskip}{-10pt}
\subsection{Effect of Adversarial Learning}
We develop an adversarial training procedure to reduce the noise in the annotated dataset produced by \textit{DA}. In Table \ref{table:finalresults}, we find that GCNN (Target) method trained on the annotated target dataset constructed by \textit{DA} achieves impressive performance on all the five datasets, outperforming the WEB-CWS method. In addition, with the adversarial training module, the model further yields the remarkable improvements of the F1-scores. The results demonstrate that the adversarial network could capture deeper semantic features than simply using the GCNN-CRF model, via better making use of the information from both source and target domains.
\begin{figure}[h]
\setlength{\belowcaptionskip}{-10pt}
\centering
\subfigure[Features on DM.]{\includegraphics[width=0.235\textwidth]{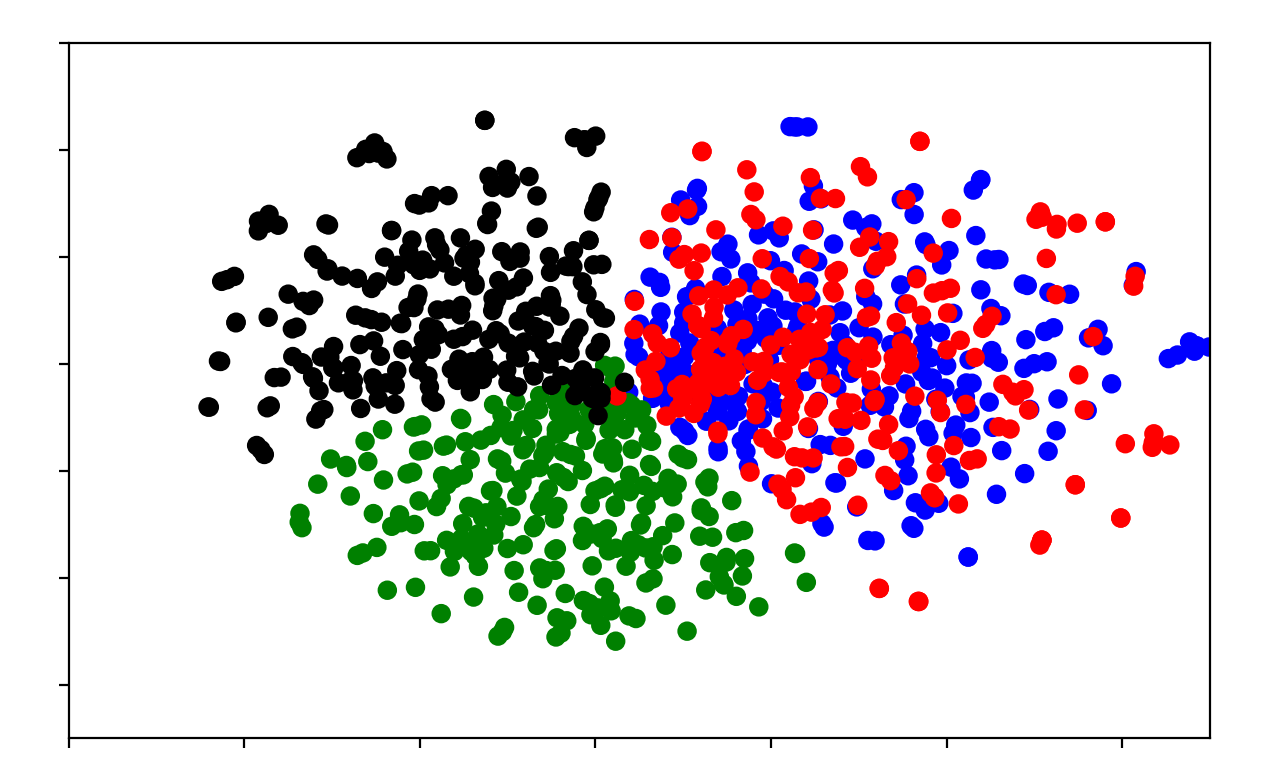}}
\subfigure[Features on DL. ]{\includegraphics[width=0.235\textwidth]{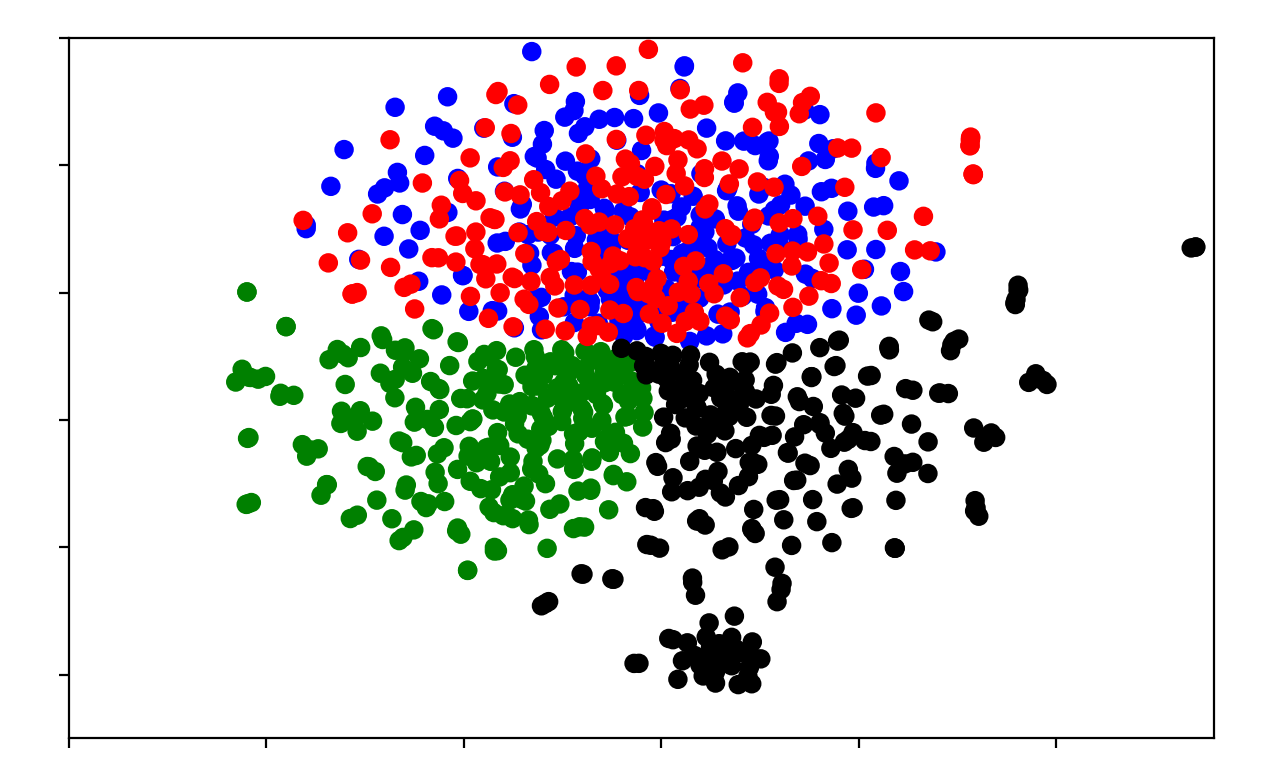}}
\caption{$t$-SNE visualisation of ${\bm H}$ and ${\bm H}^*$ produced by the domain independent encoder and sharing encoder. Where green points $\rightarrow$ ${\bm H}_s$. black points $\rightarrow$ ${\bm H}_t$, blue points $\rightarrow$ ${\bm H}_s^*$, red points $\rightarrow$ ${\bm H}_t^*$}.
\label{fig_adv}
\end{figure}

\subsection{Analysis of Feature Distribution}
As introduced in \ref{sub:adv}, in the process of adversarial learning, domain-independent encoders could learn domain-specific features $\bm{H}_s$ and $\bm{H}_t$, and the sharing encoder could learn domain-agnostic features $\bm{H}^*_s$ and $\bm{H}^*_t$. We use $t$-SNE \cite{maaten2008visualizing} algorithm to project these feature representations into planar points for visualization to further analyze the feature learning condition. As illustrated in Figure \ref{fig_adv}, domain-independent features $\bm{H}_s$ (green) and $\bm{H}_t$ (black) have little overlap, indicating the distribution gap between different domains. However, the domain-agnostic feature distributions $\bm{H}^*_s$ (red) and $\bm{H}^*_t$ (blue) are very similar, implying that the learned feature representation can be well shared by both domains.



\subsection{Impact of Amount from Source and Target data}
\begin{figure}[t]
\centering
\includegraphics[width=0.455\textwidth]{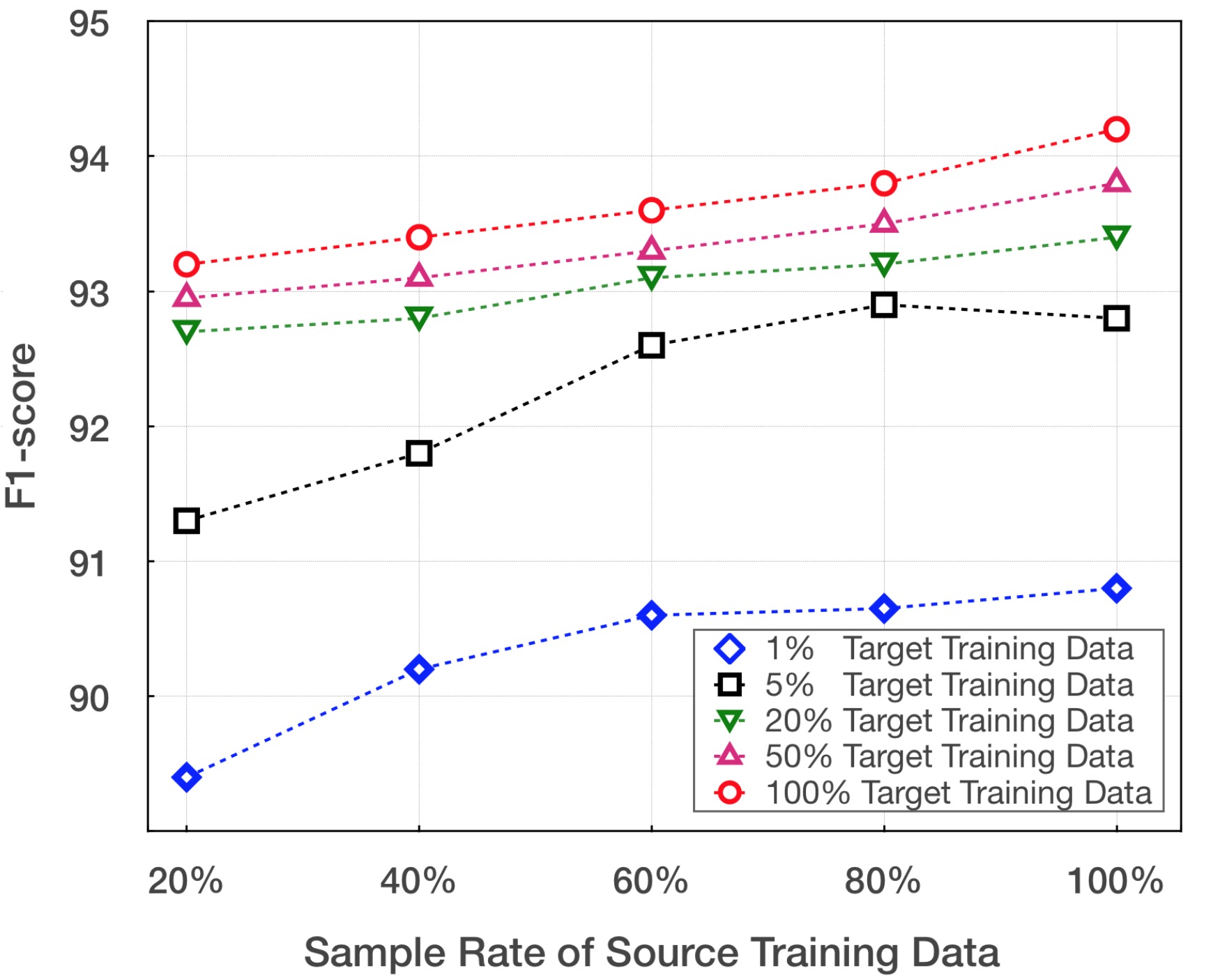} 
\caption{The impact of data amount for the source and target data on PKU (source, 47.3k sentences) and DL (target, 40.0k sentences).}
\label{fig:size}
\end{figure}

In this subsection, we analyze the impact of the data usage for both source and target domain, the experiment is conducted on the PKU (source) and DL (target) datasets. In Figure \ref{fig:size}, we respectively select $20\%$, $40\%$, $60\%$, $80\%$ and $100\%$ of the source domain data and $1\%$, $5\%$, $20\%$, $50\%$, $100\%$ of the target domain data to perform the training procedure. The result demonstrates that increasing source and target data will both lead to an increase F1-score. Generally, the amount of the target data gives more impact on the whole performance, which conforms to the intuition. The `` $1\%$ Target Training Data'' line indicates that the performance of the model will be strictly limited if the target data is severely missing. But when the amount of the target data increase to $5\%$, the performance will be improved significantly, which shows the ability to explore domain-specific information for our method. 

\section{Conclusion}
\setlength{\belowcaptionskip}{-10pt}
In this paper, we intuitively propose a unified framework via coupling distant annotation and adversarial training for the cross-domain CWS task. In our method, we investigate an automatic distant annotator to build the labeled target domain dataset, effectively address the OOV issue. Further, an adversarial training procedure is designed to capture information from both the source and target domains. Empirical results show that our framework significantly outperforms other proposed methods, achieving the state-of-the-art result on all five datasets across different domains.

\section*{Acknowledgments}

We sincerely thank all the reviewers for their insightful comments and suggestions. This research is partially supported by National Natural Science Foundation of China (Grant No. 61773229 and 61972219), the Basic Research Fund of Shenzhen City (Grand No. JCYJ20190813165003837), and Overseas Cooperation Research Fund of Graduate School at Shenzhen, Tsinghua University (Grant No. HW2018002).

\bibliography{acl2020}
\bibliographystyle{acl_natbib}
\end{document}